\documentclass[journal]{IEEEtran}
\usepackage{graphicx}
\usepackage{subfig}
\usepackage{caption}
\usepackage{multirow}
\usepackage{float}
\usepackage{algorithm}  
\usepackage{algorithmic} 
\usepackage{amsmath}
\usepackage{amssymb}
\usepackage{makecell}
\usepackage{color}
\usepackage{colortbl}

\usepackage{xcolor}
\definecolor{rblue}{rgb}{0,0.5,1}
\usepackage[bookmarks=false]{hyperref}
\hypersetup{colorlinks=true,linkcolor=red,citecolor=rblue}

\begin{document}
\title{PVPUFormer: Probabilistic Visual Prompt Unified Transformer for Interactive Image Segmentation}

\author{Xu~Zhang,
Kailun~Yang\IEEEauthorrefmark{2},
Jiacheng~Lin,
Jin~Yuan\IEEEauthorrefmark{1}\IEEEauthorrefmark{2},
Zhiyong~Li\IEEEauthorrefmark{1},~\IEEEmembership{Member,~IEEE},
and Shutao~Li,~\IEEEmembership{Fellow,~IEEE}
\thanks{X. Zhang, J. Lin, J. Yuan, and Z. Li are with the College of Computer Science and Electronic Engineering, Hunan University, Changsha 410082, China.}
\thanks{K. Yang, Z. Li, and S. Li are with the School of Robotics and the National Engineering Research Center of Robot Visual Perception and Control Technology, Hunan University, Changsha 410082, China.}
\thanks{S. Li is also with the College of Electrical and Information Engineering and with the Key Laboratory of Visual Perception and Artificial Intelligence of Hunan Province, Hunan University, Changsha 410082, China.}
\thanks{\IEEEauthorrefmark{1}Corresponding authors: Jin Yuan and Zhiyong Li. (E-mail: yuanjin@hnu.edu.cn, zhiyong.li@hnu.edu.cn.)}
\thanks{\IEEEauthorrefmark{2}Equal advising.}
}

\markboth{IEEE Transactions on Image Processing, October~2024}
{Zhang \MakeLowercase{\textit{et al.}}: VPUFormer}

\maketitle

\begin{abstract}
Integration of diverse visual prompts like clicks, scribbles, and boxes in interactive image segmentation significantly facilitates users' interaction as well as improves interaction efficiency. However, existing studies primarily encode the position or pixel regions of prompts without considering the contextual areas around them, resulting in insufficient prompt feedback, which is not conducive to performance acceleration. To tackle this problem, this paper proposes a simple yet effective Probabilistic Visual Prompt Unified Transformer (PVPUFormer) for interactive image segmentation, which allows users to flexibly input diverse visual prompts with the probabilistic prompt encoding and feature post-processing to excavate sufficient and robust prompt features for performance boosting. Specifically, we first propose a Probabilistic Prompt-unified Encoder (PPuE) to generate a unified one-dimensional vector by exploring both prompt and non-prompt contextual information, offering richer feedback cues to accelerate performance improvement. On this basis, we further present a Prompt-to-Pixel Contrastive (P$^2$C) loss to accurately align both prompt and pixel features, bridging the representation gap between them to offer consistent feature representations for mask prediction. Moreover, our approach designs a Dual-cross Merging Attention (DMA) module to implement bidirectional feature interaction between image and prompt features, generating notable features for performance improvement. A comprehensive variety of experiments on several challenging datasets demonstrates that the proposed components achieve consistent improvements, yielding state-of-the-art interactive segmentation performance. Our code is available at \url{https://github.com/XuZhang1211/PVPUFormer}. 
\end{abstract}

\begin{IEEEkeywords}
Interactive image segmentation, Transformer, Visual prompt, Contrastive loss.
\end{IEEEkeywords}

\IEEEpeerreviewmaketitle

\section{Introduction} \label{introduction}
\IEEEPARstart{I}{mage} segmentation, which aims to partition an input image into meaningful parts~\cite{yang2021context,wang2018probabilistic,li2020personal,ding2022deep_matting}, has sparked enthusiasm in computing vision due to its wide spread of applications in automatic driving \cite{li2020deep}, robots \cite{reddy2022first}, and et.  
Benefiting from the significant progress of deep learning, existing image segmentation methods have undergone a rapid performance leap, but still cannot accurately segment desired targets at one time. Consequently, interactive image segmentation, which aims to complement defective segmentation results in an image by iteratively inputting prompts like scribbles~\cite{Errortolerant_bai2014error, ScribbleSeg_chen2023scribbleseg}, clicks~\cite{DIS_xu2016deep, RITM_sofiiuk2022reviving,tang2022active, lin2023adaptiveclick}, and boxes~\cite{boxprior_lempitsky2009image, GrabCut_rother2004grabcut, Milcut_wu2014milcut,tang2021look}, has attracted increasing attention, recently. The interactive feedback between a system and users could help the system accurately capture users' intentions as well as improve its algorithms to yield promising segmentation results to meet users' requirements.

\begin{figure}[!t]
\centering
\includegraphics[width=1.0\linewidth]{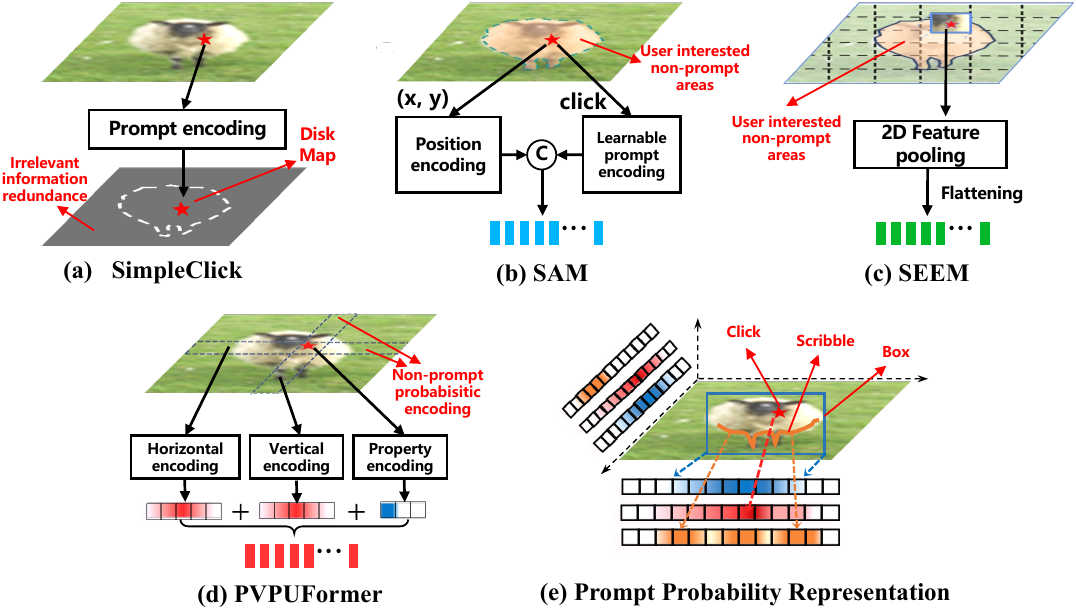}
\caption{Comparison of different prompt encoding strategies, where the two-dimensional prompt encoding (subfigure (a)) introduces irrelevant information, the one-dimensional prompt encoding (subfigure (b) and (c)) ignores contextual regions usually of interest to users. Our prompt encoding (subfigure (d)) adopts a probabilistic estimation way to encode both prompt and non-prompt information and could convert clicks, boxes and scribbles into a unified probability representation (see subfigure (e), the darker the color, the higher the probability), offering richer feedback cues for performance boosting.}
\label{fig:mot}
\end{figure}

Early interactive segmentation methods~\cite{boxprior_lempitsky2009image, Errortolerant_bai2014error,protiere2007interactive} primarily receive a single type of visual prompt during interaction to update segmentation results, significantly constraining users' behaviors as well as diminishing the efficiency of interactive segmentation.
Generally, different visual prompts have different advantages.
Click-based prompts are quick and efficient but provide limited information, leading to low segmentation precision. In contrast, scribble prompts provide rich information but are time-consuming and less efficient, while box prompts serve as a middle ground to allow users to obtain an approximate boundary of the target area in relatively less time. In the initial stages of interaction, users tend to employ click-based prompts to obtain a coarse segmentation result by considering labeling costs. Then, it is effective to use box or scribble prompts for fine-grained corrections and adjustments. Therefore, 
allowing a variety of visual prompt inputs is conducive to improving interaction efficiency and could offer a flexible interactive interface for users.
Motivated by this, the recently proposed SAM integrates a variety of prompts including clicks, boxes, masks, and text to guide image segmentation, while SEEM employs a unified visual sampler to convert all kinds of non-textual prompts to visual representations that are lying in the same visual embedding space. In contrast to the two-dimensional prompt encoding strategies ~\cite{DIS_xu2016deep, LIS_benenson2019large} containing irrelevant information redundancy on non-prompt regions (see Fig.~\ref{fig:mot} (a)), the proposed encoders in SEEM~\cite{seem_zou2023segment} and SAM~\cite{SAM_kirillov2023segment} adopt one-dimensional encoding for visual prompts, either on their positions (see Fig.~\ref{fig:mot} (b)) or region features (see Fig.~\ref{fig:mot} (c)), which significantly enhances interactive efficiency, but still encounters a critical issue stemming from its binary encoding strategy, that is, it only encodes prompt pixels and discard non-prompt regions during interactive process. This binary encoding strategy only explores limited confident prompt information, usually resulting in slow performance improvement. The surrounding regions around a visual prompt are usually also of interest to a user (see Fig.~\ref{fig:mot} (b) and (c)), and the use of these non-prompt regions could help the system to better guess users' intention for performance acceleration. Unfortunately, existing prompt encoding fails to consider this contextual non-prompt information. 

Towards this end, this paper proposes a Probabilistic Visual Prompt Unified Transformer (PVPUFormer) for Interactive Image Segmentation, which integrates multiple types of visual prompts including clicks, boxes, and scribbles in a unified probabilistic representation format. 
Considering text input is typically utilized in automatic referring image segmentation instead of interactive segmentation, our framework does not consider textual prompts during the interactive process. Instead, this study focuses on effective visual prompt encoding and post-processing to accelerate performance improvement. Specifically, we first propose a Probabilistic Prompt-unified Encoder (PPuE) to unify all types of visual prompts in a one-dimensional probabilistic vector concatenated by a horizontal representation vector, a vertical representation vector, and an intention property vector as shown in Fig.~\ref{fig:mot} (d). 
The horizontal/vertical probabilistic representation vector is calculated according to the spatial and visual distances between a prompt pixel and a non-prompt pixel, where the smaller distance between them indicates a higher probability of the non-prompt pixel having the same intention proper as the prompt.
As shown in Fig. 1 (e), all three types of prompts can be converted into a unified horizontal/vertical probabilistic encoding based on both spatial distance and visual similarity. The probabilistic value gradually decreases around the click across the whole image width and height, while that value directly reduces to zero outside the boundary of the box. For the scribble, since it contains multiple positive clicks, the probabilistic distribution presents multiple high peaks.
Different from one-dimensional prompt encoding in SAM and SEEM, our prompt encoding adopts a probabilistic representation vector to sufficiently excavate contextual non-prompt regions around prompts, thereby offering richer non-prompt information for performance improvement.
On this basis, our approach further performs post-processing on encoded prompts from two aspects: 
First, we introduce a Prompt-to-Pixel Contrastive (P$^2$C) loss to perform feature alignment between
prompt features and pixel features. Initially, we transform probabilistic prompt representations into visual feature representations using MLP mapping. Subsequently, the P$^2$C loss calculates the similarity between prompt features and pixel features, aiming to pull close them with the same label and push away them with different labels, effectively bridging the representation gap between prompt and pixel representations for the model's optimization. To the best of our knowledge, this is the first attempt to align prompt features and pixel features for interactive image segmentation.
Second, we design a Dual-cross Merging Attention (DMA) module to implement bidirectional feature interaction. The prompt-to-semantic cross-attention selectively extracts image features guided by prompt features, which could filter irrelevant image regions. Meanwhile, the semantic-to-prompt cross-attention helps improve prompt representations, yielding better prompt features for the model's updating. 

We extensively evaluate our method on several public benchmarks, and the experimental results demonstrate that the proposed components are all effective, enabling PVPUFormer to yield state-of-the-art performance as compared to existing interactive image segmentation methods. At a glance, the main contributions are summarized as follows:

\begin{itemize}
\item We propose an effective Probabilistic Visual Prompt Unified Transformer (PVPUFormer) for interactive image segmentation. Beyond existing prompt encoding strategies, the proposed Probabilistic Visual Prompt Encoder (PPuE) considers both prompt and non-prompt regions in a probabilistic estimation way, offering richer feedback information to accelerate performance improvement.
\item We are the first to employ a Prompt-to-Pixel Contrastive (P$^2$C) loss for interactive image segmentation, which effectively bridges the representation gap between pixel and prompt features, thereby offering consistent feature representations to support accurate mask prediction. 
\item We design a Dual-cross Merging Attention (DMA) module to implement bidirectional feature interaction, which could extract notable prompt and image features as well as effectively filter irrelevant ones, thereby enhancing the accuracy of mask prediction. 
\end{itemize}

\begin{figure*}[ht!]
	\centering
	\includegraphics[width=1.0\linewidth]{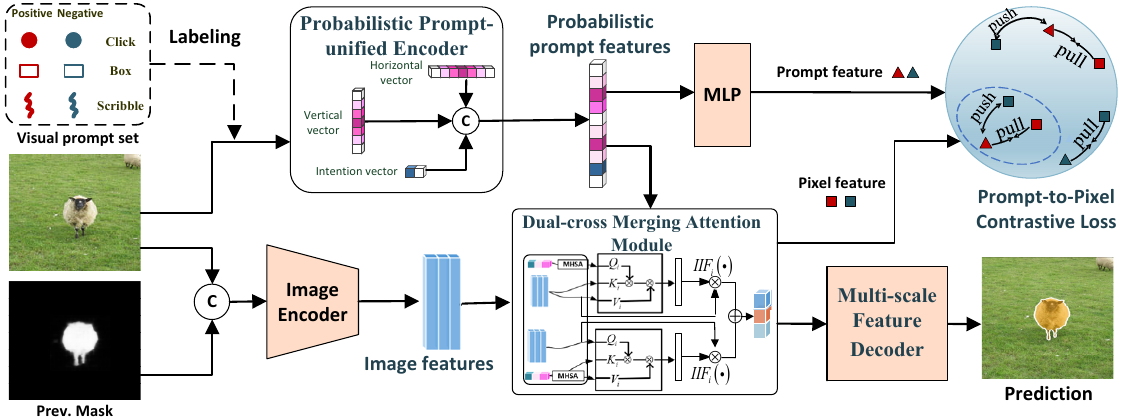}
	\caption{The pipeline of the proposed Probabilistic 
 Visual Prompt Unified Transformer (PVPUFormer), which consists of four components: a Probabilistic Prompt-unified Encoder (PPuE), an Image Encoder, a Dual-cross Merging Attention (DMA) module, and a Multi-scale Feature Decoder.}
	\label{fig:networks}
\end{figure*}
\section{Related Work}  \label{related}
\subsection{Interactive Image Segmentation} 
Early interactive image segmentation approaches mainly adopt optimization-based methods~\cite{Randomwalks_grady2006random} to minimize a specifically constructed cost function defined on a graph over image pixels~\cite{GrabCut_rother2004grabcut, graphcut_boykov2001interactive, Loosecut_yu2017loosecut}.
Thanks to the advance of deep learning, recent studies have developed a variety of deep learning models for interactive image segmentation~\cite{BRS_jang2019interactive, f_brs_sofiiuk2020f, 99accuracy_forte2020getting}.
For instance, Xu~\textit{et al.}~\cite{DIS_xu2016deep} first introduced a deep model to transform positive and negative clicks into separate Euclidean Distance Maps, and then concatenates the maps with an input image as a composite input to a Convolutional Neural Network (CNN) for mask prediction. RITM~\cite{RITM_sofiiuk2022reviving} extends click-based interactive segmentation to allow modifying existing instance segmentation masks interactively, which has inspired numerous subsequent research works in this field. 
GPCIS~\cite{GPCIS_zhou2023interactive} formulates the click-based interactive segmentation task as a pixel-wise binary classification model based on Gaussian processes (GP). It employs amortized variational inference to approximate the GP posterior in a data-driven way and then decouples the approximated GP posterior into dual-space forms for efficient sampling with linear complexity. Besides CNN, transformer-based models have been also employed for interactive image segmentation~\cite{f_brs_sofiiuk2020f, IStransformer_faizov2022interactive}, where a user's clicks are still concatenated with an image as an input to the models for mask prediction. Benefiting from the self-attention mechanism, transformer-based approaches have demonstrated promising performance for interactive image segmentation. This work also adopts transformer-based backbones for interactive image segmentation.
Differently, our model supports multiple types of visual prompts and focuses on developing an effective probabilistic prompt encoding and post-processing to boost segmentation performance. 

\subsection{Different Types of Interactive Feedback}
Most interactive image segmentation approaches adopt click prompts as 
users' feedback for its simplicity and efficiency \cite{FocalClick_chen2022focalclick, FCFI_wei2023focused}. However, since click prompts have a limited receptive field, various works have been devoted to exploring other prompts for interactive feedback.  For example, \cite{boxprior_lempitsky2009image} and \cite{Milcut_wu2014milcut} adopt bounding boxes as feedback queries, which can effectively define the range of a desired region but face uncertainty in region boundaries when dealing with irregular contours.
Zhang~\textit{et al.}~\cite{zhang2020interactive} utilized an inside point near the center of an object and two outside points at the symmetrical corners of a tight bounding box to address this limitation, generating extra labeling costs. Besides bounding boxes \cite{boxprior_lempitsky2009image, GrabCut_rother2004grabcut, Milcut_wu2014milcut}, scribbles~\cite{Errortolerant_bai2014error, ScribbleSeg_chen2023scribbleseg} are also used for interactive image segmentation, which could provide rich and precise information to capture users' intention but requires users to invest more time and knowledge as compared to boxes and clicks.
Apart from employing a single form of prompts, several works \cite{Phraseclick_ding2020phraseclick, SAM_kirillov2023segment, seem_zou2023segment} have explored employing a combination of various forms of prompts for interactive segmentation.
For instance, Kirillov~\textit{et al.}~\cite{SAM_kirillov2023segment} leveraged learnable vectors with position embeddings to represent various types of prompts.
Zou~\textit{et al.}~\cite{seem_zou2023segment} constructed a promptable, interactive universal segmentation model, where a visual sampler is used to extract prompt points including clicks, boxes, and scribbles with the corresponding point feature vectors as a user's feedback. Although unified representations of various visual prompts have achieved promising performance, the above methods only focus on utilizing labeled visual prompts to capture users' intentions, which offers limited feedback information for performance acceleration.
In summary, click-based prompts are quick and efficient but provide limited information. In contrast, scribble prompts provide rich information but are time-consuming and less efficient, while box prompts serve as a middle ground to allow users to obtain an approximate boundary of the target area in relatively less time.
Our approach considers encoding both prompt and non-prompt areas in a probabilistic estimation way. It integrates multiple types of visual prompts including clicks, boxes, and scribbles into a unified probabilistic representation, providing richer feedback cues to enhance performance.
Different from PPL~\cite{kwon2023probabilistic}, which utilizes probabilistic prompts by learning them from the semantic information in both images and text to capture class attributes, our approach directly converts user prompts into a unified probabilistic prompt encoding, enabling more effective capture of user intent.

\subsection{Iterative Optimization for Local Details}
Recent approaches~\cite{FocalClick_chen2022focalclick, FCFI_wei2023focused} focus on local refinement for interactive image segmentation due to its efficiency and effectiveness. Compared to global refinement, local refinement aims at exploring the differences between the current prediction and the previous prediction.
For instance, FocalClick~\cite{FocalClick_chen2022focalclick} efficiently updates the mask in the region that the user intends to modify and retains predictions in other regions.
FocusCut~\cite{Focuscut_lin2022focuscut} integrates the functions of object segmentation and local refinement. After obtaining the global prediction, it crops click-centered patches from the original image with adaptive scopes to refine the local predictions progressively.
FCFI~\cite{FCFI_wei2023focused} focuses on a local area around the new click and subsequently corrects the feedback based on the similarities of high-level features.
It alternately updates and collaboratively refines the feedback and deep features to integrate the feedback into the features. 
Differently, our approach aims to align prompt representations and pixel representations in contrastive learning for the model's optimization, which could bridge the huge representation gap between them as well as yield robust visual features for mask prediction.

\section{VPUFormer: Proposed Architecture}  \label{Approach}

\subsection{Overview}  \label{method:overview}

Fig.~\ref{fig:networks} illustrates the architecture of our proposed Probabilistic Visual Prompt Unified Transformer (PVPUFormer), which consists of four main components: a Probabilistic Prompt-unified Encoder (PPuE), an image encoder, a Dual-cross Merging Attention (DMA) module, and a multi-scale feature decoder.
Specifically, given an image labeled with visual prompts to indicate desired (positive) or irrelevant (negative) regions by users, we first employ the PPuE to convert multiple types of visual prompts into a unified probabilistic representation, as well as use the image decoder to extract the visual feature of the image, respectively.
Then, we inject both image and prompt representations into our Dual-cross Merging Attention (DMA) module, which implements bidirectional feature interaction between them to generate notable and noiseless visual features for mask prediction.
Finally, the multi-scale feature decoder upsamples the multi-scale features via the feature pyramid network structure,  and then predicts a probability map for mask prediction. To bridge the representation gap between prompt and image representations, we propose a Prompt-to-Pixel Contrastive (P$^2$C) loss, which could effectively pull close the corresponding prompt and image representations with the same label as well as push away non-matching ones with different labels, yielding consistent feature representations to support effective mask prediction.

Different from the previous studies, our PVPUFormer first adopts a Probabilistic Prompt-unified Encoder to encode both prompt and non-prompt information, offering richer feedback cues to accelerate performance improvement. Moreover, the proposed P$^2$C loss and DMA module could effectively align both image and prompt features as well as explore notable visual features, respectively, offering robust visual features to support accurate mask prediction.

\subsection{Probabilistic Prompt-unified Encoder} \label{method:QUE}

Beyond existing prompt encoding strategies \cite{ScribbleSeg_chen2023scribbleseg, RITM_sofiiuk2022reviving, GrabCut_rother2004grabcut}, the proposed Probabilistic Prompt-unified Encoder (PPuE) simultaneously considers both prompt and non-prompt visual cues, and adopts a probabilistic estimation way to offer richer feedback information to accelerate performance improvement.  

\begin{figure*}[t!]
	\centering
\includegraphics[width=1.0\linewidth]{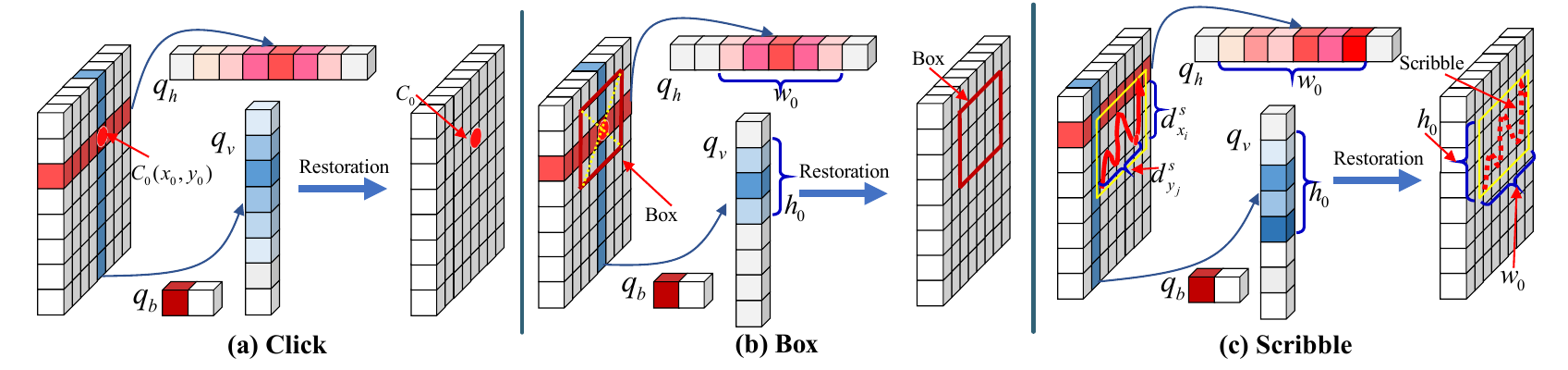}
	\caption{Three examples to show the click, box, and scribble encoding by the PPuE, respectively, where the PPuE constructs a one-dimension prompt vector $q$ to represent a visual prompt, composing of three parts: a horizontal representation vector $q_h$, a vertical representation vector $q_v$, and an intention property vector $q_b$.}
	\label{fig:pue}
\end{figure*}

Fig.~\ref{fig:pue} illustrates the encoding of clicks, boxes, and scribbles by using the PPuE.
To effectively capture a user's intention, the PPuE constructs a one-dimension prompt vector $q$ to represent the encoding result, which is composed of three parts as shown in Fig.~\ref{fig:pue} (a): a horizontal representation vector $q_h$, a vertical representation vector $q_v$, and an intention property vector $q_b$.
The intention property vector records the ``positive'' (inside the desired mask) or ``negative'' (outside the desired mask) property of a prompt, while the horizontal/vertical representation vector indicates the property probability distribution in the horizontal/vertical direction for a given image according to the prompt.
Next, we elaborate on how to encode clicks, boxes, and scribbles, respectively.

\textbf{Click Encoding.} Given a positive/negative click $C_{0}(x_0, y_0)$ on an image $\boldsymbol{I} \in \mathbb{R}^{H \times W \times 3}$, where $H$ and $W$ are the height and width of $\boldsymbol{I}$, and $(x_0, y_0)$ is the click's coordinate.
Click encoding aims at generating a horizontal representation vector $\boldsymbol{q}_{h} \in \mathbb{R}^W$ and a vertical representation vector $\boldsymbol{q}_{v} \in \mathbb{R}^H$, which reflect the property probability distribution in the horizontal and vertical directions.
Taking horizontal representation vector generation as an example, two assumptions are made according to the spatial  and visual distances for the property probability estimation:
First, if a point has a close spatial distance to the click in the horizontal direction, the probability of them having the same property is high;
Second, if a point has a close pixel value as the click, indicating the similar visual appearances between them, then that probability is also high.

Based on the assumptions, given a point $C_i(x_i, y_i)$ in $q_h$, we first calculate the spatial distance $d_{x_i}^{s}$ and the visual distance $d_{x_i}^{v}$ between them as follows:
\begin{equation}
    d_{x_i}^{s}=\sqrt{\left ( x_i-x_0 \right )^2 } , i\in \left [ 0,W \right)
\label{sdistance}
\end{equation}
\begin{equation}
    d_{x_i}^{v}=\sqrt{\left (p_{x_i}-p_{x_0} \right )^2 } , i\in \left [ 0,W \right ) , \label{vdistance}
\end{equation}
where $p_{x_i}$ denotes the pixel value of $C_i(x_i, y_i)$. We then multiply them as the final distance $d_{x_i}$, which generates a distance vector $D_h$ in the horizontal direction. On this basis, we employ Quasi-Gaussian \cite{gaussianvector_xiong2020gaussian} with a standard deviation $\sigma$ to convert $D_h$ to a horizontal representation vector $q_{h}$ as follows:
\begin{equation}
    {q}_{h}^{i}=\left\{\begin{array}{ll}
    e^{-\frac{{d_{x_i}}^2}{2 \sigma^{2}}}, & \text { if } d_{x_i} \leq \sigma \\
    0, & \text { otherwise, }
    \end{array}\right. 
\label{click}
\end{equation}
where ${q}_{h}^{i}$ is the $i$-th element in $q_{h}$.
In the same way, we can obtain $\boldsymbol{q}_{v}$, and the final click encoding vector $\boldsymbol{q}_{click}$ is generated by concatenating $\boldsymbol{q}_{h}$, $\boldsymbol{q}_{v}$, and $\boldsymbol{q}_{b}$ as follows:
\begin{equation}
   \boldsymbol{q}_{click}=[\boldsymbol{q}_{h},\boldsymbol{q}_{v},\boldsymbol{q}_{b}],
\end{equation}
where $[,]$ is the concatenation operation, and $\boldsymbol{q}_{b}$ is the one-hot encoding result of the property ``positive'' or ``negative''. 

For all the elements in $\boldsymbol{q}_{click}$, there are only two elements assigned with the property probability $1$, where the first element reflects the horizontal position by $C_0(x_0, y_0)$, and the second indicates the vertical position.
Thus, the representation vector $\boldsymbol{q}_{click}$ records the location of a user's click as well as the property probability of non-prompt areas.
Moreover, compared to 2D sparse representations like Disk Map~\cite{RITM_sofiiuk2022reviving}, our approach requires less storage and well reflects the property probability distribution for mask prediction.

\textbf{Box Encoding.}
Similar to click encoding, box encoding aims at generating horizontal and vertical representation vectors $\boldsymbol{q}_{h} \in \mathbb{R}^W$ and $\boldsymbol{q}_{v} \in \mathbb{R}^H$ to reflect the property probability distribution in the horizontal and vertical directions given a box prompt $B_0(x_0, y_0, w_0, h_0)$, where $(x_0, y_0)$ is its center coordinates, and $(w_0, h_0)$ is its width and height.
We assume that the center point $(x_0, y_0)$ has the highest probability of satisfying the input property, and a point with a closer distance would have a higher property probability, which is the same as the click encoding.
Differently, a box prompt gives the boundary information, which explicitly indicates that the points outside the boundary violate the prompt property. Therefore, we revise Eq.~(\ref{sdistance}) as follows:  
\begin{equation}
 d_{x_i}^{s}=\left\{\begin{array}{ll}
    \sqrt{\left ( x_i-x_0 \right )^2 } & \text { if }  |x_i-x_0| \in \left [ 0,\frac{w_0}{2} \right) \\
    + \infty, & \text { otherwise. }
    \end{array}\right.
\end{equation}
As a result, the element ${q}_{h}^{i}$ in $q_{h}$ outside the box boundary would be assigned with zero in Eq.~(\ref{click}).
Compared to click encoding, box encoding offers precise boundary information, yielding a better prompt representation vector for mask prediction. 

\textbf{Scribble Encoding.} 
Given a scribble prompt $S(C_1, ..., C_N)$, where $C_1, ..., C_N$ denote the points on the scribble $S$, and the point $C_n$ is located in the position $(x_n,y_n)$, we assume the intersection point between $\boldsymbol{q}_{h} \in \mathbb{R}^W$ and $\boldsymbol{q}_{v} \in \mathbb{R}^H$ (see Fig.~\ref{fig:pue} (c)) is located at the top-left corner of the scribble bounding box, and aim to estimate the property probability of each element in $\boldsymbol{q}_{h}$/$\boldsymbol{q}_{v}$. Similarly, if a point in $\boldsymbol{q}_{h}$/$\boldsymbol{q}_{v}$ has a closer distance to the scribble, it has a higher property probability as same as the scribble property. However, different from clicks and boxes, a scribble is usually an irregular curve composed of continuous points, whose number greatly exceeds the number of elements in $\boldsymbol{q}_{h}$ and $\boldsymbol{q}_{v}$, posing great challenges for scribble encoding. 

To tackle this issue, we adopt an approximate strategy to discretize the continuous scribble into a finite number of points $m = (w_0 + h_0)$, where $w_0$ and $h_0$ represent the width and height of the bounding box of the scribble. Concretely, as shown in Fig.~\ref{fig:pue} (c), given a point in $\boldsymbol{q}_{h}$ or $\boldsymbol{q}_{v}$, our approach first randomly selects one of the aligned points from the scribble as the candidate. Here, an aligned point has the same horizontal/vertical coordinate with the point in $\boldsymbol{q}_{h}$/$\boldsymbol{q}_{v}$. Then, we adopt the click encoding strategy to calculate the distance between the point and the candidate, and then convert the distance into a property probability. 

Different from click and box encoding, scribble encoding only records partial points on a scribble to approximately preserve its contour information, resulting in a certain amount of information loss. Nonetheless, scribble encoding offers sufficient information to capture users' intention to improve segmentation results. As shown in Algorithm 1, taking the generation of $\boldsymbol{q}_{h}$ as an example, given a point $C_i(x_i,y_{i})$ in $q_h$ and a vertical alignment point $C_n(x_n, y_{n})$ on the scribble, we can calculate the distance $d_{x_i}^s$ between them as follows:
\begin{equation}
 d_{x_i}^{s}=\left\{\begin{array}{ll}
 \sqrt{\left(y_{i}-y_{n}\right)^{2}}, & \text { if } x_i \in B(x_0,y_0,w_0,h_0)  \\
    + \infty, & \text { otherwise.}
    \end{array}\right.
\end{equation}
where $B(x_0,y_0,w_0,h_0)$ is the bounding box of the scribble $S$ centered in $(x_0,y_0)$ with width $w_0$, height $h_0$. Following click encoding, we then convert the distance vector $D_h$ into a probability distribution according to Eq.~(\ref{click}). Although scribble encoding suffers from a certain information loss, it could preserve the contour information of a scribble to accurately capture a user's intention.

\begin{algorithm}[h!]
\caption{Scribble Encoding}
\begin{algorithmic}[1]
\REQUIRE Scribble $S(C_1, ... ,C_N)$, where $C_1, ... ,C_N$ are points on the scribble;
\STATE Set box $B(x_0,y_0,w_0,h_0)$ as the bounding box of the scribble, where $(x_0,y_0)$ is its center point, and $w_0,h_0$ are its width and height, respectively;
\STATE Initialization: horizontal/vertical vector $\boldsymbol{q}_h \in \mathbb{R}^W$, $\boldsymbol{q}_v \in \mathbb{R}^H$, and the one-hot encoding property $\boldsymbol{q}_b$, a standard deviation $\sigma$ for the Quasi-Gaussian;

\FOR {each point $C_i(x_i,y_i)$ in $\boldsymbol{q}_h$}
\IF{$x_i$ $\notin$ $B(x_0,y_0,w_0,h_0)$}
\STATE $q_h^i=0$
\ELSE
\STATE $C_n(x_n, y_n) \gets$ randomly select a point from $S(C_1, ..., C_N)$, where $x_n=x_i$,

\STATE $d_{x_{i}}^{s}=\sqrt{\left(y_{i}-y_{n}\right)^{2}}$,
\STATE ${q}_{h}^{i}=e^{-\frac{{d_{x_i}^{s}}^2}{2 \sigma^{2}}} \text { if } d_{x_i}^{s} \leq \sigma, \text { else } {q}_{h}^{i}=0 $
\STATE remove $C_n(x_n,y_n)$ from $S$.
\ENDIF
\ENDFOR

\FOR {each point $C_j(x_j,y_j)$ in $\boldsymbol{q}_v$}
\IF{$y_j$ $\notin$ $B(x_0,y_0,w_0,h_0)$}
\STATE $q_v^j=0$
\ELSE
\STATE $C_n(x_n, y_n) \gets$ randomly select a point from $S(C_1, ..., C_N)$, where $y_n=y_j$,
\STATE $d_{y_{j}}^{s}=\sqrt{\left(x_{j}-x_{n}\right)^{2}}$,
\STATE ${q}_{v}^{j}=e^{-\frac{{d_{y_j}^{s}}^2}{2 \sigma^{2}}} \text { if } d_{y_j}^{s} \leq \sigma, \text { else } {q}_{v}^{j}=0 $

\ENDIF
\ENDFOR
\STATE $\boldsymbol{q}_{scrbble} \gets \text { concatenate }      \boldsymbol{q}_{h},\boldsymbol{q}_{v},\boldsymbol{q}_{b}$

\ENSURE $\boldsymbol{q}_{scribble}$
\end{algorithmic}
\end{algorithm}

In summary, the proposed PPuE allows users to flexibly input visual prompts, and efficiently integrates both valuable prompt and non-prompt visual cues for interactive image segmentation, offering concise and rich feedback information for mask prediction.

\subsection{Dual-cross Merging Attention} \label{method:MDM}
Dual-cross Merging Attention (DMA) aims to select informative visual features that exhibit the highest mutual response between a visual prompt and image features, which consists of a multi-head self-attention layer, two multi-head cross-modal attention layers, two feed-forward neural layers, and an interactive information filtering layer.

Concretely, given a prompt encoding vector $q \in \mathbb{R}^{M \times D}$ and a visual feature $f_{v} \in \mathbb{R}^{\frac{H}{16} \times \frac{W}{16} \times D}$ of an input image, where $M$ denotes the number of user interactions, and $D$ is the feature dimension, DMA first passes $q$ into a Multi-Head Self-Attention (MHSA) layer to obtain the attention prompt feature $q'\in \mathbb{R}^{M \times D}$, which highlights the important areas in an image. On this basis, the Multi-Head Cross-modal Attention (MHCA) layer~\cite{transformer_vaswani2017attention} performs the bidirectional cross-modal attention on $q'$ and $f_{v}$ to generate the prompt-to-semantic feature $F_{qv} \in \mathbb{R}^{\frac{H}{16} \times \frac{W}{16} \times D}$ and the semantic-to-prompt feature $F_{vq}\in \mathbb{R}^{M \times D}$, respectively:
\begin{equation}
    \begin{matrix}
    F_{qv}=\mathrm{MHCA}({q}^{\prime}, f_v, f_v) + f_v. \\
    F_{vq}=\mathrm{MHCA}(f_v, {q}^{\prime}, {q}^{\prime}) + q.
    \end{matrix}
\end{equation}
The prompt-to-semantic feature $F_{qv}$ explores notable visual features guided by prompt features, which could effectively filter irrelevant image regions. Comparatively, the semantic-to-prompt feature $F_{vq}$ utilizes visual features to improve prompt features, yielding accurate probabilistic prompt representations to capture users' intentions. These features are then passed through two Feed-Forward Neural (FFN) layers: 
\begin{equation}
    \begin{matrix}
    \hat{F}_{qv}=\mathrm{FFN}(\mathrm{LN}(F_{qv})) ,  \\
    \hat{F}_{vq}=\mathrm{FFN}(\mathrm{LN}(F_{vq})) ,
    \end{matrix}
\end{equation}
where LN$(\cdot)$ denotes layer normalization. Subsequently, they are fed into an Information Filtering (IF) layer to calculate the feature response and select the features with the highest response value for each prompt channel.
In detail, we use the Sigmoid function to obtain the interactive weights, and
these weights are then element-wise multiplied with the visual features to obtain $\hat{F}_{iqv} \in \mathbb{R}^{\frac{H}{16} \times \frac{W}{16} \times D}$ and $\hat{F}_{iqv} \in \mathbb{R}^{\frac{H}{16} \times \frac{W}{16} \times D}$ respectively, which helps select effective interactive information based on user prompts as well as filter out invalid and redundant information: 
\begin{equation}
    \begin{matrix}
    \hat{F}_{iqv}=\mathrm{IF}(\hat{F}_{qv},f_{v})= \mathrm{Sigmoid} (\phi (\hat{F}_{qv}))\otimes f_{v} , \\
    \hat{F}_{ivq}=\mathrm{IF}(\hat{F}_{vq},f_{v})= \mathrm{Sigmoid}(\phi (\hat{F}_{vq}))\otimes f_{v} ,
    \end{matrix}
\end{equation}
where $\phi (\cdot)$ is an operation that selects the highest interactive response value from $\hat{F}_{qv}$ or $\hat{F}_{vq}$. Finally, the bidirectional interaction feature $F_{dual}$ is formalized as follows:
\begin{equation}
F_{dual}=\hat{F}_{iqv} + \hat{F}_{ivq} .
\end{equation}

For implementation, we use three Dual-Cross Merging Attention (DMA) layers and add the positional encodings~\cite{SAM_kirillov2023segment} to multi-scale visual features. The bidirectional interaction between prompt and image representations yields noiseless and notable visual features, thereby supporting accurate mask prediction.
Unlike DM-Fusion~\cite{xu2023dm}, which primarily enhances feature complementarity across modalities, our DMA module focuses on selecting relevant interactive information based on probabilistic encoding while filtering out invalid and redundant information. This refinement significantly improves the accuracy of the interactive representation.

\subsection{Multi-scale Feature Decoder}
To capture rich multi-scale spatial information, we adopt a feature pyramid network in~\cite{FPN_lin2017feature} to combine features from different scales.
Concretely, we first use two transposed convolutional layers to upsample the bidirectional interactive features $F_{dual}$, obtaining visual features with the $1/4$, $1/16$, $1/32$, and $1/64$ size of the original image, respectively.
Subsequently, the multi-scale features are transformed to have an identical channel dimension through a $1 \times 1$ convolutional layer and then upsampled with the same resolution for concatenation, yielding a robust visual feature $\hat{F}_{v}$ for mask prediction.
Finally, the concatenated feature $\hat{F}_{v}$ is passed through an MLP layer followed by a sigmoid function to output a single-channel prediction result $O_{mlp}$, which represents a segmentation probability map for mask generation.

\subsection{Prompt-to-Pixel Contrastive (P$^2$C) Loss} \label{method:LF}

Although visual prompts can reflect user requirements to some extent, there still exists a significant difference in representation between the encoding of visual prompts and that of image vision, which greatly affects the performance of mask prediction.
To tackle this issue, we design a prompt-to-pixel contrastive loss, which explicitly aligns visual prompt features and the corresponding pixel features. Concretely, we first adopt a Multi-layer Perceptron (MLP) to map the probabilistic prompt feature $q' \in \mathbb{R}^{M \times D}$ into a visual prompt feature, followed by the scale normalization on both image feature $\hat{F}_{v}$ and prompt feature $q'$ as follows:
\begin{equation}
    \begin{matrix}
        z_v = \mathrm{normalize}(\hat{F}_{v}), \\
        z_q = \mathrm{normalize}(MLP(q')),
    \end{matrix}
\end{equation}
where $z_v \in \mathbb{R}^{\frac{H}{4} \times \frac{W}{4} \times D}$, $z_q \in \mathbb{R}^{M \times D}$ are the representations of image and prompt in the new space. Next, we calculate the similarity $\rho \in \mathbb{R}^{M \times \frac{H}{4} \times \frac{W}{4}}$ between $z_q$ and $z_v$ through a dot product operation as follows:
\begin{equation}
\label{dot_product}
\rho = \frac{1}{2}(z_{q} \cdot z_{v}^\top  + 1).
\end{equation}
Each element $\rho_{i,j}$ in $\rho$ reflects the representation similarity between the $i$-th prompt in $z_q$ and the $j$-th pixel in $z_v$. It is expected that the similarity value $\rho_{i,j}$ is $1$ when the $j$-th pixel belongs to the mask indicated by the $i$-th prompt, otherwise $0$. As a result, we design a Prompt-to-Pixel (P$^2$C) loss, which is calculated as follows:  
\begin{equation}
\begin{aligned}
\mathbb{\ell }_{\text{P}^2\text{C}} (z_{q}^{i}, z_{v}^{j}) & =\left\{\begin{array}{ll}
-\log (\rho_{i,j}), & Y_{i,j} \in \mathcal{P}, \\
-\log (1 - \rho_{i,j}), &otherwise,
\end{array}\right. 
\end{aligned}
\end{equation}
where $\mathcal{P}$ denotes a set of the matching (prompt, pixel) pairs, and  $Y_{i,j}$ represents a pair of the $i$-th prompt and the $j$-th pixel. Finally, the $\text{P}^2\text{C}$ loss function is expressed as:
\begin{equation}
\mathbb{\ell }_{\text{P}^2\text{C}} =\frac{1}{M\times L} \sum_{i=0}^{M-1}\sum_{j=0}^{L-1}\mathbb{\ell }_{\text{P}^2\text{C}} (z_{q}^{i}, z_{v}^{j}),
\end{equation}
where $L=\frac{H}{4} \times \frac{W}{4}$ is the flatten length.
The $\text{P}^2\text{C}$ loss well pulls closer the representations of prompts and the corresponding pixel features, as well as pushes away the non-matching pairs, promoting our model to learn robust features for prompts and images to bridge the representation gap between them. As a result, the prompt could better help predict the desired mask based on consistent feature representations.

 On this basis, our approach integrates three losses including the weighted cumulative NFL loss~\cite{RITM_sofiiuk2022reviving, sun2024cfr}, the DICE loss~\cite{dice_milletari2016v}, and the proposed P$^2$C loss to train the model, which is expressed as: 
\begin{equation}
  \mathbb{L }_{\mathrm{total}} = \mathbb{\ell }_{\mathrm{NFL}} + \mathbb{\ell }_{\text{ DICE}} + \lambda \mathbb{\ell }_{\text{P}^2\text{C}}, \label{totalloss}
\end{equation}
where $\lambda$ is a hyperparameter to adjust the scale of $\mathbb{\ell }_{\text{P}^2\text{C}}$.

\section{Experiments} \label{Experiment}
In this section, we first introduce our datasets and experimental settings, followed by the illustration of experimental results with detailed analysis.

\begin{table*}[t!]
\caption{Evaluation results tested on GrabCut, Berkeley, SBD, and DAVIS datasets where our model is trained on the SBD dataset. Throughout this paper, the best and second-best results are denoted in \textbf{bold} and \underline{underlined}, respectively.}
\renewcommand\arraystretch{1.1}
\centering
\resizebox{\linewidth}{!}{
\begin{tabular}{l|l|l|cc|cc|cc|cc}
\hline
\multirow{2}{*}{Method} & \multirow{2}{*}{Backbone}  & \multirow{2}{*}{Train Data}  & \multicolumn{2}{c|}{GrabCut} & \multicolumn{2}{c|}{Berkeley} & \multicolumn{2}{c|}{SBD} & \multicolumn{2}{c}{DAVIS} \\ \cline{4-11} 
 &  & & NoC@85 & NoC@90 & NoC@85 & NoC@90 & NoC@85 & NoC@90 & NoC@85 & NoC@90 \\ \hline
RITM \cite{RITM_sofiiuk2022reviving} & \multirow{4}{*}{SegFormerB0-S2} & \multirow{4}{*}{SBD} & 1.62 & 1.82 & \underline{1.84} & 2.92 & 4.26 & 6.38 & 4.65 & 6.13 \\
FocalClick \cite{FocalClick_chen2022focalclick} &  &  & 1.66 & 1.90 & - & 3.14 & 4.34 & 6.51 & 5.02 & 7.06 \\
GPCIS \cite{GPCIS_zhou2023interactive} &  &  & \underline{1.60} & \underline{1.76} & \textbf{1.84} & \underline{2.7} & \underline{4.16} & \underline{6.28} & \underline{4.45} & \underline{6.04} \\
\textbf{PVPUFormer} & &  & \textbf{1.54} & \textbf{1.68} & 1.87 & \textbf{2.53} & \textbf{4.10} & \textbf{5.96} & \textbf{4.24} & \textbf{5.78} \\ \hline
f-BRS-B \cite{f_brs_sofiiuk2020f} & \multirow{7}{*}{ResNet50} & \multirow{7}{*}{SBD} & 2.20 & 2.64 & 2.17 & 4.22 & 4.55 & 7.45 & 5.44 & 7.81 \\
CDNet \cite{CDNet_chen2021conditional} &  &  & 2.22 & 2.64 & - & 3.69 & 4.37 & 7.87 & 5.17 & 6.66 \\
RITM \cite{RITM_sofiiuk2022reviving} &  &  & 2.16 & 2.3 & 1.9 & 2.95 & 3.97 & 5.92 & 4.56 & 6.05 \\
FocusCut \cite{Focuscut_lin2022focuscut} &  &  & \underline{1.60} & \textbf{1.78} & 1.86 & 3.44 & 3.62 & \underline{5.66} & 5 & 6.38 \\
FocalClick \cite{FocalClick_chen2022focalclick} &  &  & 1.92 & 2.14 & 1.87 & 2.86 & 3.84 & 5.82 & 4.61 & 6.01 \\
GPCIS \cite{GPCIS_zhou2023interactive} &  &  & 1.64 & \underline{1.82} & \underline{1.60} & \underline{2.60} & \underline{3.80} & 5.71 & \underline{4.37} & \underline{5.89} \\
\textbf{PVPUFormer} & &   & \textbf{1.58} & 1.86 & \textbf{1.52} & \textbf{2.39} & \textbf{3.72} & \textbf{5.60} & \textbf{3.94} & \textbf{5.64} \\ \hline
RITM \cite{RITM_sofiiuk2022reviving} & \multirow{4}{*}{HRNet-18s} & \multirow{4}{*}{SBD} & 2.00 & 2.24 & 2.13 & 3.19 & 4.29 & 6.36 & 4.89 & 6.54 \\
FocalClick \cite{FocalClick_chen2022focalclick} &  &  & 1.86 & 2.06 & - & 3.14 & 4.3 & 6.52 & 4.92 & 6.48 \\
GPCIS \cite{GPCIS_zhou2023interactive} &  &  & \underline{1.74} & \underline{1.94} & \underline{1.83} & \textbf{2.65} & \underline{4.28} & \underline{6.25} & \textbf{4.62} & \underline{6.16} \\
\textbf{PVPUFormer} & &   & \textbf{1.65} & \textbf{1.82} & \textbf{1.80} & \underline{2.68} & \textbf{4.12} & \textbf{5.87} & \underline{4.75} & \textbf{6.13} \\ \hline
\end{tabular}
}
\label{compare_sota_sbd}
\end{table*}

\begin{table*}[t!]
\caption{Evaluation results on GrabCut, Berkeley, SBD, DAVIS, COCO MVal, and ADE20K datasets, where our model is trained on the COCO + LIVS or SA-1B dataset.}
\renewcommand\arraystretch{1.1}
\centering
\tabcolsep=0.05cm 
\resizebox{\linewidth}{!}{
\begin{tabular}{l|l|l|cc|c|cc|cc|cc|cc}
\hline
\multirow{2}{*}{Method} & \multirow{2}{*}{Backbone} & \multirow{2}{*}{Train Data} & \multicolumn{2}{c|}{GrabCut} & \multicolumn{1}{c|}{Berkeley} & \multicolumn{2}{c|}{SBD} & \multicolumn{2}{c|}{DAVIS} & \multicolumn{2}{c|}{COCO MVal} & \multicolumn{2}{c}{ADE20K} \\ \cline{4-14} 
 &  & & NoC@85 & NoC@90 & NoC@90 & NoC@85 & NoC@90 & NoC@85 & NoC@90 & NoC@85 & NoC@90 & NoC@85 & NoC@90 \\ \hline

f-BRS-B \cite{f_brs_sofiiuk2020f} & HRNet32 & \multirow{7}{*}{COCO-LVIS} & 1.54 & 1.69 & 2.44 & 4.37 & 7.26 & 5.17 & 6.50 & 2.35 & 3.44 & - & - \\
FocalClick \cite{FocalClick_chen2022focalclick} & HRNet32 &  & 1.64 & 1.80 & 2.36 & 4.24 & 6.51 & 4.01 & 5.39 & 2.62 & 3.65 & 9.09 & 12.24 \\
DynaMITe \cite{rana2023dynamite} & HRNet32 & & 1.62 & 1.68 & \underline{2.04} & \underline{3.83} & 6.35 & \underline{3.83} & 5.2 & 2.35 & 3.14 & - & - \\

RITM \cite{RITM_sofiiuk2022reviving} & HRNet-18s &  & 1.54 & 1.68 & 2.60 & 4.26 & 6.86 & 4.79 & 6.00 & 2.40 & 3.35 & 8.37 & 11.77 \\
FCFI \cite{FCFI_wei2023focused} & HRNet-18s & & 1.50 & \textbf{1.56} &  2.05 & {3.88} & \underline{6.24} & \textbf{3.70} & \underline{5.16} & \underline{2.20} & \underline{3.04} & \underline{8.26} & \underline{11.73} \\

FocalClick \cite{FocalClick_chen2022focalclick} & HRNet-18s &  & \underline{1.48} & 1.62 & 2.66 & 4.43 & 6.79 & 3.90 & 5.25 & 2.61 & 3.59 & 9.91 & 12.93 \\

\textbf{PVPUFormer} & HRNet-18s & & \textbf{1.46} & \underline{1.59} & \textbf{1.94} & \textbf{3.76} & \textbf{6.12} & 3.91 & \textbf{5.08} & \textbf{2.18} & \textbf{2.97} & \textbf{8.20} & \textbf{11.65}  \\ \hline

DynaMITe \cite{rana2023dynamite} & SegFormerB0 & COCO-LVIS & 1.48 & 1.58 & 1.97 & 3.81 & 6.38 & 3.81 & 5.00 & 2.47 & 3.28 & - & - \\
FocalClick \cite{FocalClick_chen2022focalclick} & SegFormerB3 & COCO-LVIS& 1.44 & 1.50 & 1.92 & 3.53 & 5.59 & {3.61} & {4.90}  & 2.32 & 3.12 & 8.97 & 12.03 \\
EMC-Click \cite{du2023efficient} & SegFormerB3 & COCO-LVIS & 1.42 & 1.48 & 2.35 & 3.44 & 5.57 & 4.49 & 5.69 & \underline{2.13} & \underline{2.85} & 10.83 & 13.63 \\
FDRN \cite{zeng2023feature} & SegFormerB3 & COCO-LVIS & 1.42 & 1.44 & 1.80 & 3.74 & 5.57 & 3.55 & 4.90 & - & - & - & - \\
VTMR \cite{fang2023variance} & SegFormerB3 & COCO-LVIS & 1.38 & \underline{1.42} & \underline{1.72} & 3.55 & \underline{5.53} & \textbf{3.26} & \underline{4.82} & - & - & - & - \\

 SAM \cite{SAM_kirillov2023segment}          & ViT-B  & SA-1B & 2.42    & 2.72        & 2.96           & $6.50$   & $9.76$          & 6.13   & 7.89    & 5.70 & 8.99 & 13.40 & 16.40   \\
 SEEM \cite{seem_zou2023segment}  & DaViT-B & COCO-LVIS& -    & -        & -           & $6.67$   & $9.99$          & -   & -     & - & -  & - & -   \\
InterFormer   \cite{huang2023interformer} & ViT-B & COCO-LVIS& $1.38$    & $1.50$        & $3.14$           & $3.78$   & $6.34$          & $4.10$   & $6.19$  & - & - & - & - \\	
SimpleClick \cite{SimpleClick_liu2022simpleclick} & ViT-B & COCO-LVIS& \underline{1.38} & {1.48} & 1.97 & \underline{3.43} & 5.62 & 3.66 & 5.06 & 2.16 & 2.92 & \underline{8.32} & \underline{11.59} \\
\textbf{PVPUFormer} & ViT-B & COCO-LVIS& \textbf{1.34} & \textbf{1.40} & \textbf{1.71} & \textbf{3.32} & \textbf{5.45} & \underline{3.48} & \textbf{4.82} & \textbf{2.12} & \textbf{2.85} & \textbf{7.59} & \textbf{10.90} \\ \hline
\end{tabular}
}
\label{compare_sota_coco_lvi}
\end{table*}

\subsection{Datasets}
We trained our model on two public datasets, and tested the performance on nine testing sets including six natural datasets and three medical datasets. 

\textbf{Training Sets.} We use the following two training datasets.
\begin{itemize}
\item  SBD~\cite{SBD_hariharan2011semantic}: This dataset contains $8,498$ images for training, which is widely used as a training dataset for the interactive image segmentation task.

\item  COCO \cite{coco_lin2014microsoft}+LVIS \cite{Lvis_gupta2019lvis}:
COCO contains $118K$ training images with a total of $1.2M$ instances, and LVIS shares the same images with COCO but has more instance masks and higher mask quality.
\end{itemize}

\textbf{Testing Sets.}
We use the following testing datasets to evaluate our model.
\begin{itemize}
\item GrabCut~\cite{GrabCut_rother2004grabcut}: It contains $50$ images with $50$ instances, and each image has clear foreground and background differences. 
\item Berkeley~\cite{Berkeley_martin2001database}: This dataset includes $96$ images with $100$ instances in the validation set, which is used for evaluation in our experiments.
\item SBD~\cite{SBD_hariharan2011semantic}: This dataset contains $2,857$ validation images with $6,671$ instances. Following~\cite{RITM_sofiiuk2022reviving, FocalClick_chen2022focalclick, Focuscut_lin2022focuscut}, we evaluate our model on the validation dataset.
\item DAVIS \cite{DAVIS_perazzi2016benchmark}: This dataset contains $50$ videos, and we only use the same $345$ frames as used in \cite{f_brs_sofiiuk2020f, FocalClick_chen2022focalclick, Focuscut_lin2022focuscut, lin2020interactive} for evaluation. 
\item COCO MVal \cite{coco_lin2014microsoft}: This dataset is a subset of COCO with a total of 800 images, and contains 10 objects from each object category. 

\item ADE20K \cite{zhou2017scene}: This dataset comprises 20,210 images in the training set, 2,000 images in the validation set, and 3,000 images in the testing set. All images are meticulously annotated with objects.

\item ssTEM~\cite{ssTEM_gerhard2013segmented}: It includes two image stacks, and each contains $20$ medical images. We evaluate our model on the same stack as used in~\cite{PseudoClick_liu2022pseudoclick} for evaluation. 
\item BraTS~\cite{BraTS_baid2021rsna}: This dataset includes $369$ Magnetic Resonance Image (MRI) volumes, and we use the same $369$ slices as used in~\cite{PseudoClick_liu2022pseudoclick}. 
\item OAIZIB~\cite{OAIZIB_ambellan2019automated}: This dataset contains $507$ MRI volumes, and we test on the same $150$ slices with $300$ instances as used in~\cite{PseudoClick_liu2022pseudoclick}.
\end{itemize}

\textbf{Evaluation Metrics.}
To ensure a fair performance comparison with existing methods, we evaluate our model using the standard Number of Clicks (NoC) metric when only click prompts are used as inputs. The NoC measures the number of clicks required to achieve a predefined Intersection over the Union (IoU) threshold between predicted and ground truth masks.
We set the IoU threshold to $85\%$ and $90\%$ as NoC@85 and NoC@90, respectively.
The maximum number of clicks for each instance is set to $20$.
When multiple types of prompts (clicks, boxes, or scribbles) are used as inputs, we employ the Number of Interactions (NoI) metric, which is similar to NoC. Since it is only allowed to input one visual prompt in each interaction, NoI is equaling the number of input prompts. The Number of Failures (NoF) is also reported and it counts the number of images that cannot achieve the target IoU within $20$ clicks.
Besides, we use the average IoU to evaluate the segmentation quality given $k$ clicks (IoU@k).

\subsection{Implementation Details}
\textbf{Model settings:} To demonstrate the generality of our method, we conduct experiments on four backbones including ViT-B~\cite{vit_dosovitskiy2020image}, SegFormerB0-S2~\cite{SegFormer_xie2021segformer}, HRNet18s~\cite{hrnet_sun2019high}, and DeepLabV3+~\cite{deeplabv3_chen2018encoder} with ResNet50~\cite{resnet_he2016deep}.
For encoding, all the input images are first unified to the size of $448 {\times} 448$, and then fed to a backbone above to extract visual features. Data augmentation techniques, including random resizing (scale ranges from $0.75$ to $1.25$), random flipping and rotation, random brightness contrast, and random cropping, are used to boost performance. 
All the visual prompts are encoded into a Gaussian vector with $\sigma {=} 3$ by the PPuE, generating a $899$-dimensional vector concatenated by two $448$-dimensional horizontal and vertical vectors, and one $3$-dimensional property vector. The feature dimension $D$ of both image and prompt in the DMA module is set to $768$, which generates three bidirectional interaction features with different scales for mask prediction by the multi-scale feature decoder. For the loss function in Eq.\ref{totalloss}, we set $\lambda$ to $2$ for the model's optimization. Additionally, we input the previous forward-pass predicted mask $\boldsymbol{M} \in \mathbb{R}^{1 \times H \times W}$ to the model. Following the previous works~\cite{RITM_sofiiuk2022reviving, SimpleClick_liu2022simpleclick}, we employ a Conv1S network architecture to fuse the predicted mask and image.

\textbf{Training settings:} To train our model, the initial learning rate is $5{\times}10^{-4}$ for SegFormerB0-S2, ResNet50, and HRNet18s, and $5{\times}10^{-5}$ for ViT-B.
The learning rate is then reduced by $0.1$ after $50$ epochs. The Normalized Focal Loss (NFL)~\cite{RITM_sofiiuk2022reviving} is used during training with $\alpha {=} 0.5$ and $\gamma {=} 2$. We train our model for $55$ epochs by using the Adam optimizer ($\beta_1 {=} 0.9$ and $\beta_2 {=} 0.999$) with a batch size of $32$.
All of our models are trained on two NVIDIA RTX A6000 GPUs.

\textbf{Iterative Labeling Strategy:}
By simulating a user's habit, the system first automatically labels a visual prompt, and then the model updates the parameters to predict a mask. This process repeats until the performance exceeds the predefined IoU value or the maximum number of prompt inputs arrives. Specifically, the system first labels a positive click on a fixed position of a ground truth mask to predict the initial mask. 
Next, the system compares the predicted and ground truth masks to find the largest area of segmentation errors. Then it labels a visual prompt (positive or negative) with a random position within this area for resulting updating. This strategy is widely employed in the interactive segmentation task~\cite{FocalClick_chen2022focalclick,SimpleClick_liu2022simpleclick}. Since different models generate different masks during the interaction, and thus the prompt labeling results may be different accordingly.

To train our model, we initially input a click to generate a predicted mask, and then randomly label a click, a box, or a scribble by a random function to update results. 
A weighted cumulative NFL loss~\cite{RITM_sofiiuk2022reviving, sun2024cfr} is applied to supervise the generated mask sequence across different iterative outputs.
Since different types of prompts (click, box, scribble) are converted into a unified probabilistic encoding during training, the system supports the input of various prompt types during testing to generate segmentation results.
At each iteration, a single prompt type is selected as input for result updating. The model then generates a mask, which is concatenated with the image along the channel dimension for the next iteration.
To make a fair performance comparison with the existing methods, most of which only consider click prompts measured by NOC, our approach only labels a click in each interaction. 
In addition, we also conducted a self-assessment (see Table \ref{ablation_prompt_permutation}) by introducing a box or a scribble or both during the interaction to observe performance change, and we will give the detailed labeling strategy in the experimental analysis.

\subsection{Experimental Results}

\subsubsection{Comparison with several State-of-the-Art Approaches}

\textbf{Results on natural datasets.}
Table~\ref{compare_sota_sbd} and~\ref{compare_sota_coco_lvi} present the performance comparison results between our PVPUFormer and the state-of-the-art methods on different datasets, trained on SBD and COCO+LIVS, respectively. It is demonstrated that the proposed PVPUFormer achieves promising performance across multiple datasets and different backbones, significantly reducing the number of clicks as well as the labeling burdens by users. Moreover, we discover that although versatile segmentation methods like SAM and SEEM support unified encoding and interaction by using diverse visual prompts, their interactive performance is not so good. Comparatively, our PVPUFormer adopts effective prompt encoding and post-processing for diverse visual prompts, thereby significantly improving the interactive performance, with fewer clicks to achieve the desired mask accuracy. 

\begin{figure}[t!]
	\centering
	\includegraphics[width=1.0\linewidth]{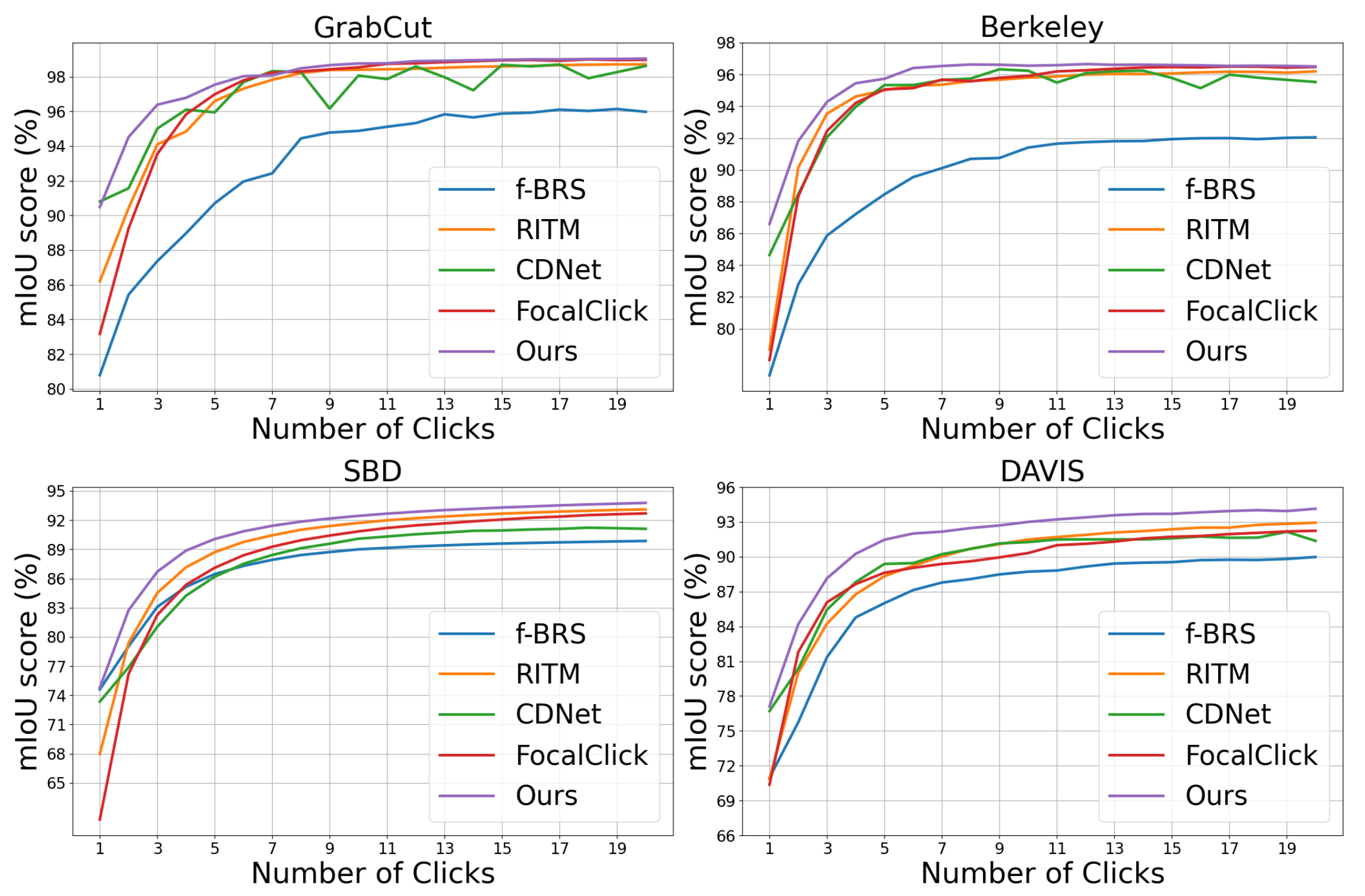}
	\caption{Comparisons of the mIoU-NoC curves on four datasets by different approaches.}
	\label{fig:miou}
\end{figure}

Fig.~\ref{fig:miou} further illustrates the mIoU-NoC line charts on four different datasets. It is demonstrated that our approach yields the best performance, with fewer clicks to achieve the same mIoU as compared to several previous methods. Specifically, our approach offers the best initial segmentation results after one click and then keeps the stable performance improvement as more clicks are labeled during the interaction. We guess that this is because our probabilistic prompt encoding could effectively capture users' intentions by exploring both prompt and non-prompt areas, thereby yielding better initial segmentation results and faster performance acceleration.

\begin{figure*}[t!]
	\centering
	\includegraphics[width=0.95\linewidth]{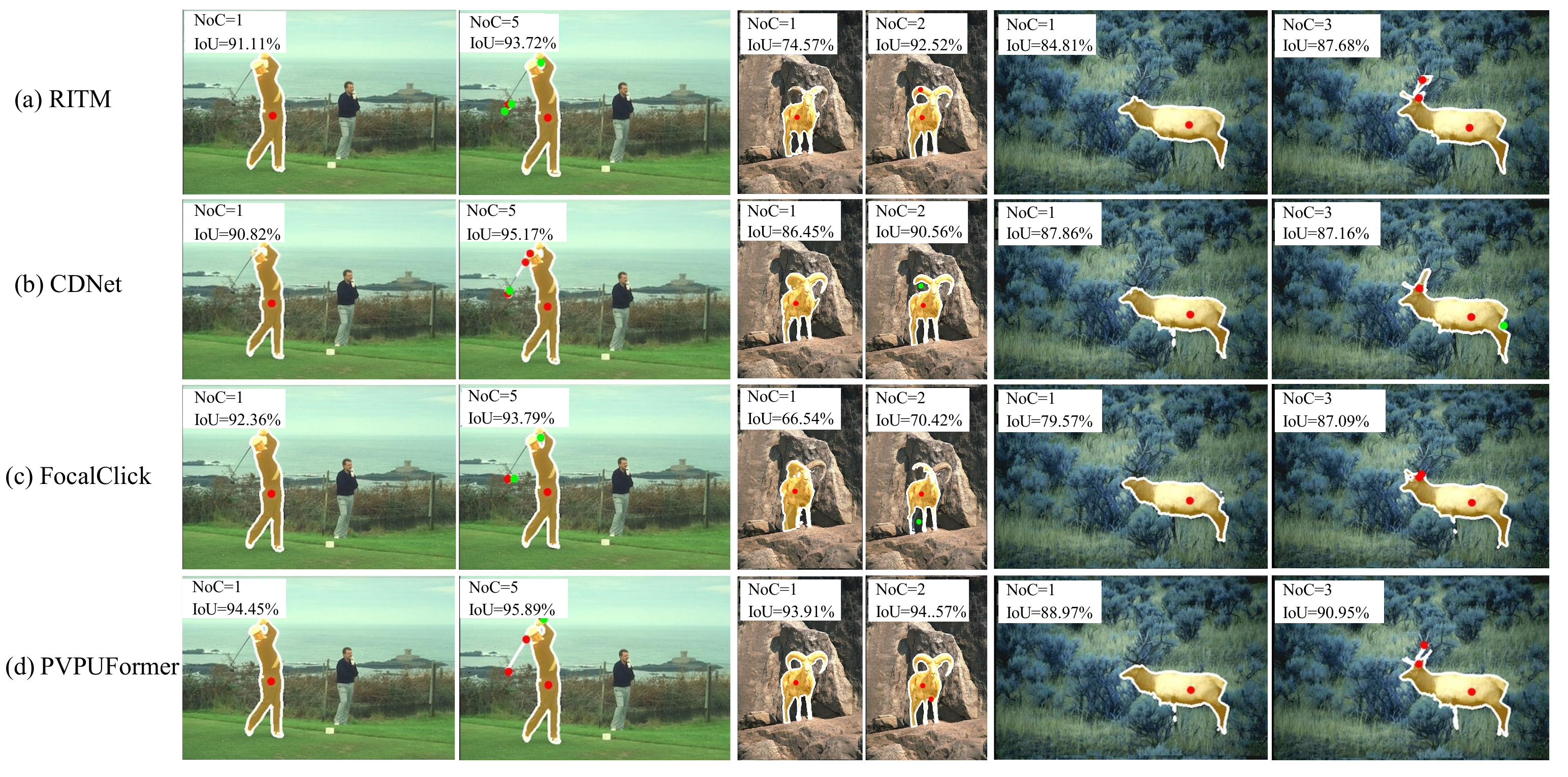}
	\caption{Qualitative comparisons of segmentation results by different approaches ( RITM~\cite{RITM_sofiiuk2022reviving}, CDNet~\cite{CDNet_chen2021conditional}, FocusClick~\cite{FocalClick_chen2022focalclick}, and our method) on three difficult examples, where the first example in the first two columns has a spidery golf clue to be masked, the second example in the middle two columns has similar foreground and background colors, and the third example has partially occluded object components.}
	\label{fig:exp1}
\end{figure*}

Fig.~\ref{fig:exp1} illustrates the quantitative results of our method and several previous methods. All the examples are first labeled with the same click, and then different approaches predict different initial results, followed by incremental clicking for mask improvement. Specifically, limited by insufficient click information in the first example (see the first column), all four methods only segment the swinging person but miss the golf club. After imposing five clicks, our method successfully captures the complete golf club, while the other methods fail to accurately predict it. When facing interference from a similar background as shown in the second example (see the third and fourth columns), our approach accurately segments the goat after labeling two clicks with an IoU value of $94.57\%$, while another method cannot well distinguish the foreground and background. For the third example (see the last two columns) with background occlusion, we discover that our method successfully segments the antlers and the front legs partially occluded by grasses after imposing the fifth click, whereas the other three methods still cannot well handle this situation.

\begin{figure*}[!t]
	\centering
	\includegraphics[width=0.93\linewidth]{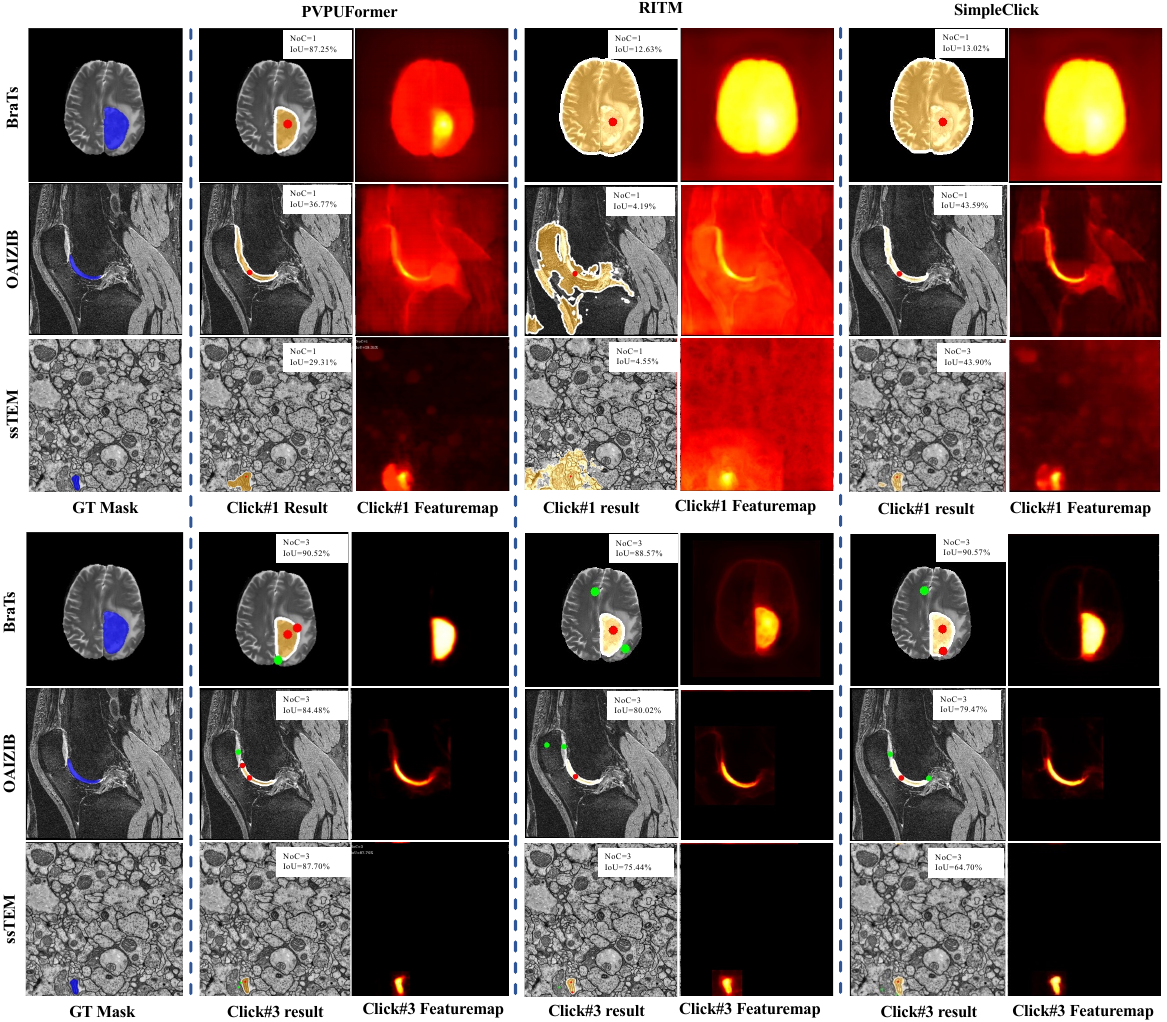}
	\caption{Three examples to visualize the segmentation results by different approaches on three medical datasets after imposing one and three clicks, respectively, where the red click represents a positive prompt, and the green click represents a negative prompt. The high brightness in the feature maps represents the large segmentation probability.}
	\label{fig:exp4}
\end{figure*}

\textbf{Results on medical datasets.}
\begin{table}[t!]
\caption{Performance comparison between PVPUFormer and several state-of-the-art methods trained on the COCO+LVIS dataset and tested on ssTEM, BraTS, and OAIZIB datasets, respectively.}
\renewcommand\arraystretch{1.4}
\setlength\tabcolsep{0.5pt}
\centering
\resizebox{\linewidth}{!}{
\begin{tabular}{l|l|cc|cc|cc}
\hline
\multirow{2}{*}{Method} & \multirow{2}{*}{Backbone} & \multicolumn{2}{c|}{ssTEM} & \multicolumn{2}{c|}{BraTS} & \multicolumn{2}{c}{OAIZIB} \\ \cline{3-8} 
 &   & NoC@85 & NoC@90 & NoC@85 & NoC@90 & NoC@85 & NoC@90 \\ \hline
CDNet \cite{CDNet_chen2021conditional}          & ResNet-34          & $4.15$   & $8.45$  & $10.51$     & $14.80$      & $17.42$   & $19.81$     \\
RITM \cite{RITM_sofiiuk2022reviving}          & HRNet32     & $\underline{2.74}$   & $\underline{4.06}$  & ${7.56}$     & $\underline{11.24}$      & $15.89$   & $19.27$  \\
RITM \cite{RITM_sofiiuk2022reviving}          & HRNet18s     & ${3.31}$   & ${4.90}$  & $\underline{7.52}$     & ${11.51}$      & $17.41$   & $19.49$  \\
FocalClick \cite{FocalClick_chen2022focalclick}      & SegF-B3             & $3.95$   & $5.05$  & $\textbf{7.17}$             & $\textbf{11.19}$      & $\textbf{12.93}$   & $19.23$          \\
SimpleClick \cite{SimpleClick_liu2022simpleclick}             & ViT-B        & $4.25$	& $5.61$	& ${8.25}$	& ${11.83}$	& $15.57$	&$\underline{18.98}$ \\
\textbf{PVPUFormer}    & ViT-B   & $\textbf{2.64}$    & $\textbf{3.90}$      & ${7.89}$      & ${11.73}$   & $\underline{14.97}$           & $\textbf{18.94}$     \\  \hline
\end{tabular}
}
\label{medical_data}
\end{table}
To evaluate the generalizability of our method, we conduct experiments on three medical image datasets as shown in 
Table~\ref{medical_data}, where we directly apply the trained models on COCO+LVIS datasets to the medical images without fine-tuning. Due to the representation gap between natural and medical images, the pre-trained models perform poorly on medical images, requiring more clicks to achieve the desired IoU as compared to that tested on natural images. We further list three qualitative results on the three medical datasets generated by PVPUFormer, RITM, and SimpleClick, respectively, as shown in Fig.~\ref{fig:exp4}. 
Obviously, our PVPUFormer could better capture a user's intention after one click, yielding more focused outcomes on both masks and feature maps, which proves the effectiveness of our encoding strategy. 
Upon further analysis, in the first row, we observe that PVPUFormer forms three distinct response regions—lesion, brain, and background—radiating from the initial click in horizontal and vertical directions. This follows our probabilistic vector model, where closer distances between prompt and non-prompt pixels indicate a higher likelihood of shared intent. In contrast, the other two methods fail to distinguish between lesion and brain regions, treating them as a single region, likely due to ineffective use of visual cues between prompt and non-prompt pixels. By imposing three clicks, PVPUFormer further improves segmentation accuracy, generating better results as compared to RTTM and SimpleClick.

\textbf{Computational Analysis.} TABLE~\ref{tab_inference_cost} provides a computation comparison between our approach and several state-of-the-art IIS methods in terms of Params (M), FLOPS (G), and inference speed (ms/c). Similar to Simpleclick~\cite{SimpleClick_liu2022simpleclick}, we evaluate the computation costs on the GrabCut dataset. Imposing new modules including PPuE and DMA, our model adds additional computational burden, with a few increases in parameters and FLOPs. Even so, the detection speed is not slow, with about $65$ ms/c to sufficiently support online feedback.
\begin{table}[t!]
\caption{Computation comparison of different models measured by Parameters (Million), FLOPS (Giga), and Speed (Millisecond per click), where * indicates the results are reproduced by us according to the provided codes by the papers. }
\centering
\renewcommand\arraystretch{1.2}
\setlength\tabcolsep{4pt}
\resizebox{\linewidth}{!}{
\begin{tabular}{l|ccc}			
\hline
 {Method (backbone, size)} &  {Params(M)} &  {FLOPs(G)} &  $ {\downarrow }$  {Speed(ms/c)} \\
\cline{1-4}

EMC-Click* (SegF-B3, 384)~\cite{du2023efficient}  & $ {45.90}$    & $ {32.3}$    & $ {152}$\\

 {RITM (HRNet32, 400)}~\cite{RITM_sofiiuk2022reviving}  & $ {30.95}$    & $ {83.12}$    & $ {54}$\\

f-BRS-B* (HRNet32, 400)~\cite{f_brs_sofiiuk2020f}  & $ {30.94}$    & $ {164.8}$    & $ {96}$\\

 FocalClick* (hrne18s, 448)~\cite{FocalClick_chen2022focalclick}  & $ {4.22}$    & $ {22.43}$    & $ {37}$\\
 FocalClick* (hrnet32, 448)~\cite{FocalClick_chen2022focalclick}  & $ {30.96}$    & $ {103.74}$    & $ {55}$\\
 FocalClick* (SegF-B3, 448)~\cite{FocalClick_chen2022focalclick}  & $ {75.78}$    & $ {24.75}$    & $ {53}$\\

 SAM* (ViT-B, 448)~\cite{SAM_kirillov2023segment} & $ {90.49}$ & $ {743.98}$ &  {88} \\
 {InterFormer (ViT-B, 512)}~\cite{huang2023interformer}  & $ {120.39}$    & $ {533.70}$    & $ {360}$\\
 {SimpleClick (ViT-B, 448)}~\cite{SimpleClick_liu2022simpleclick}  & $ {96.46}$    & $ {169.78}$    & $ {54}$\\
 %
 
% \toprule [1pt]  
\rowcolor[gray]{.9} 
 {PVPUFormer (ViT-B, 448) (ours)}  & $ {119.06}$    & $ {178.13}$    & $ {65}$\\
\hline
\end{tabular}
}

\label{tab_inference_cost}
\end{table}

\subsubsection{Evaluation on different components}
We conduct several experiments to verify the effectiveness of the proposed components including the PPuE, DMA, and P$^2$C loss.

\textbf{Evaluation on PPuE.}
This experiment verifies that PPuE can better encode visual prompts compared to the Distance map and the vector learning methods.
The Distance map represents a visual prompt as a two-dimensional map by using the Distance map method, while vector learning represents a visual prompt as a learnable embedding vector. 
From Table~\ref{ablation_prompt_type}, we can see that the use of PPuE significantly improves the performance, achieving the best NoC@85 and NoC@90 on both datasets as compared to the other two methods, with the NoC@90 $2.20$, $1.96$ on Berkeley, and $5.27$,  $5.08$ on DAVIS.
This result indicates the effectiveness of the PPuE, which produces one-dimensional Gaussian vectors to accurately capture a user's intention.

\begin{table}[t!]
\caption{Performance comparison among different prompt encoding strategies trained on COCO+LVIS dataset and tested on Berkeley~\cite{Berkeley_martin2001database} and DAVIS~\cite{DAVIS_perazzi2016benchmark} datasets.}
\centering
\renewcommand\arraystretch{1.2}
\setlength\tabcolsep{6pt}
\resizebox{\linewidth}{!}{
\begin{tabular}{l|cc|cc}
\hline
\multirow{2}{*}{\begin{tabular}[c]{@{}l@{}} Encoding Type\end{tabular}} & \multicolumn{2}{c|}{Berkeley} & \multicolumn{2}{c}{DAVIS}   \\ \cline{2-5} 
  & NoC@85 & NoC@90  & NoC@85 & NoC@90  \\ \hline
Distance map & 1.57 & 2.20  & 3.83 & 5.27  \\
Learning vector & 1.43 & 1.96  & 3.62 & 5.08  \\
\rowcolor[gray]{.9} 
PPuE vector & \textbf{1.38} & \textbf{1.71} & \textbf{3.48} & \textbf{4.82}  \\ \hline
\end{tabular}
}
\label{ablation_prompt_type}
\end{table}

\begin{table}[t!]
\caption{Performance comparison of PVPUFormer with different components tested on Berkeley~\cite{Berkeley_martin2001database} and DAVIS~\cite{DAVIS_perazzi2016benchmark} datasets.}
\centering
\renewcommand\arraystretch{1.3}
\setlength\tabcolsep{2pt}
\resizebox{\linewidth}{!}{
\begin{tabular}{c|cc|cc|cc}
\hline
\multirow{2}{*}{Backbone} & \multicolumn{2}{c|}{Method} & \multicolumn{2}{c|}{Berkeley} & \multicolumn{2}{c}{DAVIS} \\ \cline{2-7} 
 & DMA & P$^2$CL & \multicolumn{1}{c|}{NoC@90} & NoF$_{20}$@90 & \multicolumn{1}{c|}{NoC@90} & NoF$_{20}$@90 \\ \hline
\multirow{4}{*}{HRNet-18s} & - & - & \multicolumn{1}{c|}{2.53} & 6 & \multicolumn{1}{c|}{5.53} & 78 \\
 & $\checkmark$ & - & \multicolumn{1}{c|}{2.28} & 3 & \multicolumn{1}{c|}{5.25} & 61 \\
 & - & $\checkmark$ & \multicolumn{1}{c|}{2.16} & 2 & \multicolumn{1}{c|}{5.37} & 58 \\
 \rowcolor[gray]{.9} 
 & $\checkmark$ & \textbf{$\checkmark$} & \multicolumn{1}{c|}{\textbf{1.94}} & 1 & \multicolumn{1}{c|}{\textbf{5.08}} & \textbf{56} \\ \hline
\multirow{4}{*}{ViT-B} & - & - & \multicolumn{1}{c|}{2.46} & 2 & \multicolumn{1}{c|}{5.48} & 56 \\
 & $\checkmark$ & - & \multicolumn{1}{c|}{1.92} & 1 & \multicolumn{1}{c|}{5.26} & 51 \\
 & - & $\checkmark$ & \multicolumn{1}{c|}{2.13} & 1 & \multicolumn{1}{c|}{5.12} & 49 \\
 \rowcolor[gray]{.9} 
 & $\checkmark$ & \textbf{$\checkmark$} & \multicolumn{1}{c|}{\textbf{1.71}} & \textbf{0} & \multicolumn{1}{c|}{\textbf{4.82}} & \textbf{48} \\ \hline
\end{tabular}
}
\label{ablation_diff_components}
\end{table}

\textbf{Ablation study on DMA and P$^2$C loss.}
Table~\ref{ablation_diff_components} shows the performance comparison results, where ``-'' on DMA means we use the traditional Transformer to replace the DMA module. 
Compared to the baseline (``-'',``-''),
the use of DMA or P$^2$C loss significantly improves the performance. As aforementioned,
the DMA module implements effective bidirectional feature interaction to offer robust visual features for mask prediction, while the P$^2$C loss could well align both pixel and prompt features to bridge the representation gap between them.  When both modules are combined, our method achieves significant error reductions measured in NoC and NoF, which verifies the effectiveness of the proposed components.

\textbf{Evaluation on P$^2$C loss.} To investigate the impact of the P$^2$C loss on the model's performance, we adjust its weight by setting different $\lambda$ values in Eq.\ref{totalloss}. In Table~\ref{ablation_diff_hyperparameters}, the hyperparameter $\lambda$ is set in the range of $[0,5]$. We can observe that when $\lambda$ is set to $0$ (\textit{i.e.}, without using the P$^2$C loss), the performance is the worst on both datasets. As $\lambda$ continuously increases, the best results are achieved when $\lambda$ is $2$ on the DAVIS dataset, and $\lambda$ is $0.5$ or $1$ on the Berkeley dataset. This result indicates the effectiveness of our proposed P$^2$C loss, which could help learn consistent and effective prompt features for performance boosting.
The further increase of $\lambda$ leads to a performance drop since it could overshadow the effectiveness of our loss components in Eq.~\ref{totalloss}.
\begin{table}[t!]
\caption{The impact of hyperparameter settings on PVPUFormer, where we train our model on COCO+LVIS dataset~\cite{SBD_hariharan2011semantic} and test on Berkeley~\cite{Berkeley_martin2001database} and DAVIS~\cite{DAVIS_perazzi2016benchmark} datasets.}
\centering
\renewcommand\arraystretch{1.1}
\setlength\tabcolsep{9pt}
\resizebox{\linewidth}{!}{
\begin{tabular}{c|cc|cc}
\hline
\multirow{2}{*}{$\lambda$} & \multicolumn{2}{c|}{Berkeley} & \multicolumn{2}{c}{DAVIS} \\ \cline{2-5} 
 & NoC@85 & NoC@90 & NoC@85 & NoC@90 \\ \hline
0 & 1.45 & 1.92 & 3.81 & 5.26 \\
0.1 & 1.43 & 1.76 & 3.76 & 5.10 \\
0.5 & 1.41 & \textbf{1.66} & 3.94 & 5.18 \\
1 & \textbf{1.37} & 1.68 & 3.70 & 5.03 \\ 
\rowcolor[gray]{.9} 
2 & 1.38 & 1.71 & \textbf{3.48} & \textbf{4.82} \\
5 & 1.45 & 1.72 & 3.92 & 5.12 \\ \hline
\end{tabular}
}
\label{ablation_diff_hyperparameters}
\end{table}

\textbf{Evaluation on extensibility of PPuE and P$^2$C loss.}
This experiment verifies the extensibility of our proposed PPuE and P$^2$C loss. We embed the PPuE and P$^2$C loss into two representative IIS methods ``RITM'' and ``SimpleClick'', respectively to observe the performance change as shown in Table \ref{extend_module}. Obviously, the introduction of the PPuE or P$^2$C loss brings a performance increase on both approaches, which proves that they are indeed effective for the IIS task since they could offer better prompt representation to capture a user's intention, accelerating the performance improvement under limited prompt feedback.

\begin{table}[t!]
\caption{Ablation experiments of the proposed modules (PPuE, P2CL) embedded to other methods, trained on COCO+LVIS and tested on the GrabCut~\cite{GrabCut_rother2004grabcut}, Berkeley~\cite{Berkeley_martin2001database}, SBD~\cite{SBD_hariharan2011semantic}, DAVIS~\cite{DAVIS_perazzi2016benchmark} dataset.}
\centering
\renewcommand\arraystretch{1.3}
\setlength\tabcolsep{1pt}
\resizebox{\linewidth}{!}{
\begin{tabular}{c|cc|cc|c|cc|cc}
\hline
\multirow{2}{*}{Backbone} & \multicolumn{2}{c|}{Method} & \multicolumn{2}{c|}{GrabCut} & \multicolumn{1}{c|}{Berkeley} & \multicolumn{2}{c|}{SBD} & \multicolumn{2}{c}{DAVIS} \\ \cline{2-10} 
 & PPuE & P$^2$C & NoC@85 & NoC@90 & NoC@85 & NoC@85 & NoC@90 & NoC@85 & NoC@90 \\ \hline
 
\multirow{4}{*}{RITM} & - & - & 1.54 & 1.68 & 2.60 & 4.26 & 6.86 & 4.79 & 6.00 \\
 & $\checkmark$ & - & 1.52 & 1.68 & 2.57 & 4.23 & 6.68 & 4.74 & 5.88 \\
 & - & $\checkmark$ & 1.50 & 1.65 & 2.54 & 4.21 & 6.60 & 4.69 & 5.92 \\
 \rowcolor[gray]{.9} 
 & $\checkmark$ & \textbf{$\checkmark$} & \textbf{1.47} & \textbf{1.62} & \textbf{2.46} & \textbf{4.19} & \textbf{6.61} & \textbf{4.65} & \textbf{5.86} \\ \hline

\multirow{4}{*}{SimpleClick} & - & - & 1.38 & 1.48 & 1.97 & 3.43 & 5.62 & 3.66 & 5.06 \\
 & $\checkmark$ & - & 1.36 & 1.42 & 1.84 & 3.41 & 5.53 & 3.51 & 5.01 \\
 & - & $\checkmark$ & 1.30 & 1.42 & 1.80 & 3.37 & 5.56 & 3.55 & 4.98 \\
 \rowcolor[gray]{.9} 
 & $\checkmark$ & $\checkmark$ & \textbf{1.28} & \textbf{1.40} & \textbf{1.79} & \textbf{3.34} & \textbf{5.48} & \textbf{3.46} & \textbf{4.83} \\ \hline
\end{tabular}
}
\label{extend_module}
\end{table}

\subsubsection{Evaluation on the use of diverse visual prompts}
We conduct experiments to quantitatively analyze the impact of combining different types of user prompts on the model's performance as shown in Table~\ref{ablation_prompt_permutation}. 
The initial prompt is a click, and then the system randomly adopts one of the prompt candidates for feedback to update segmentation results. 
From Table~\ref{ablation_prompt_permutation}, it is seen that the performance is lowest when only clicks are used for feedback, and the introduction of boxes or scribble would significantly boost the performance. As aforementioned, a box or scribble could offer more accurate information to capture a user's intention compared to a click, thereby accelerating the performance improvement. What is more, we discover that the use of scribbles achieves better performance as compared to the use of boxes. This is because scribbles could offer accurate property information inside a box, while a box only gives a coarse indicator of a user's intention.  
When combining three types of prompts for feedback, there is a further improvement in both datasets, which indicates that using multiple types of prompts in interactive segmentation tasks is conducive to performance boosting as different types of prompts could offer richer feedback cues in complex scenarios, generating faster performance improvement as compared to the use of single prompt.
Our PPuE effectively leverages the advantages of different types of prompts by encoding them into a unified probabilistic representation.

\begin{table}[t!]
\caption{Performance comparison by using different types of visual prompts, where the models are trained on COCO+LVIS dataset and tested on  Berkeley~\cite{Berkeley_martin2001database} and DAVIS~\cite{DAVIS_perazzi2016benchmark} datasets.}
\centering
\renewcommand\arraystretch{1.1}
\setlength\tabcolsep{1pt}
\resizebox{\linewidth}{!}{
\begin{tabular}{p{1cm}<{\centering}|p{1cm}<{\centering}|p{1cm}<{\centering}|cc|cc}
\hline
\multicolumn{1}{c|}{\multirow{2}{*}{Click}} & \multicolumn{1}{c|}{\multirow{2}{*}{Box}} & \multicolumn{1}{c|}{\multirow{2}{*}{Scribble}} & \multicolumn{2}{c|}{Berkeley} & \multicolumn{2}{c}{DAVIS} \\ \cline{4-7} 
\multicolumn{1}{l|}{} & \multicolumn{1}{l|}{} & \multicolumn{1}{l|}{} & NoI@85 & NoI@90 & NoI@85 & NoI@90 \\ \hline
$\checkmark$ & - & - & 1.38 & 1.71 & 3.48 & 4.82 \\
$\checkmark$ & $\checkmark$ & - & 1.14 & 1.67 & 3.01 & 4.65 \\
$\checkmark$ & - & $\checkmark$ & 1.10 & 1.65 & 3.06 & 4.62 \\
\rowcolor[gray]{.9} 
$\checkmark$ & $\checkmark$ & $\checkmark$ & \textbf{1.10} & \textbf{1.59} & \textbf{2.94} & \textbf{4.57} \\ \hline
\end{tabular}
}
\label{ablation_prompt_permutation}
\end{table}

\begin{figure}[t!]
	\centering
	\includegraphics[width=1.0\linewidth]{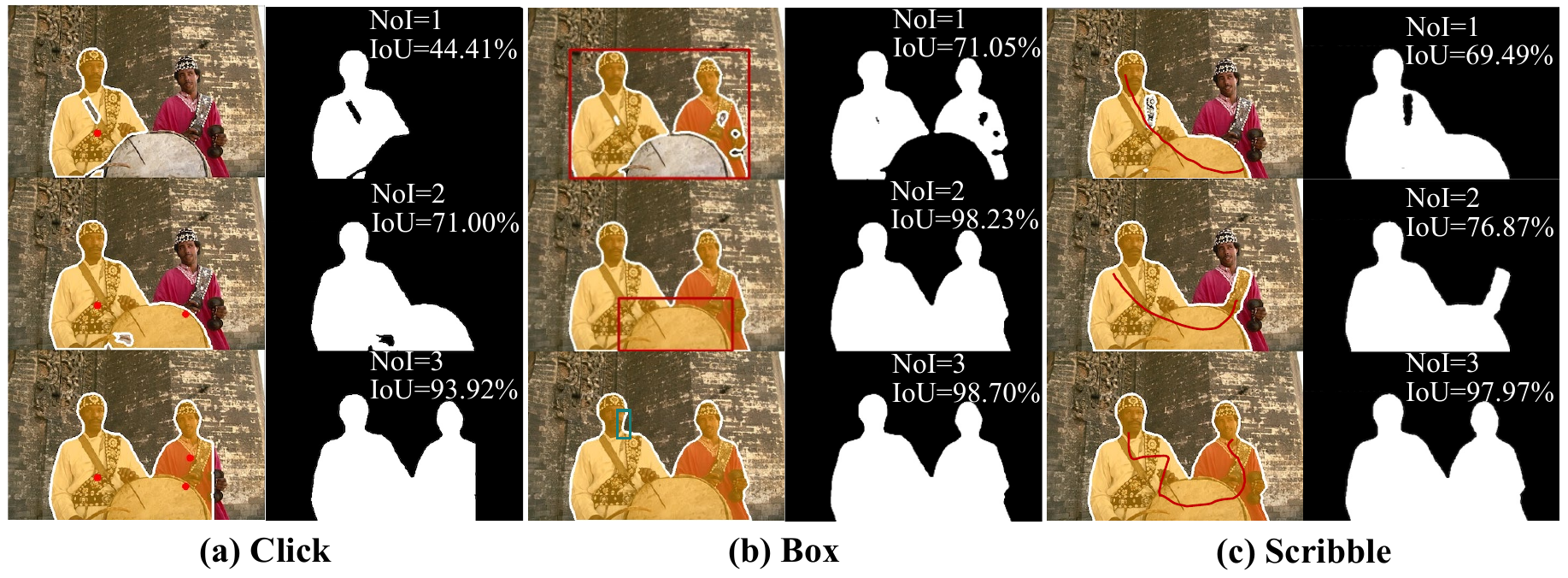}
	\caption{An example to visualize the segmentation results by using clicks, boxes, and scribbles, respectively.}
	\label{fig:exp2}
\end{figure}

Fig.~\ref{fig:exp2} lists an example to compare the interactive segmentation results by using clicks, boxes, and scribbles, respectively. It can be seen that the use of clicks has the lowest IoU values as compared to the use of boxes or scribbles, especially in the first interaction. This is because the information provided by a single click is insufficient, leading to uncertainty in the semantics to be segmented. Comparatively, the use of boxes or scribbles could provide richer feedback cues, thereby obtaining a higher IoU. Furthermore, as shown in Fig.~\ref{fig:exp2} (b), each box position is calculated based on the deviation region between the previous prediction and the ground truth mask. If the deviation region belongs to the foreground, the box is considered as a positive prompt (red box), otherwise a negative one. This strategy can correct error areas as quickly as possible for performance improvement.

\subsection{Limitations and Future Perspectives}
Despite the advantages of PVPUFormer, it still has the following limitations:
Firstly, it only considers unified prompt encoding for clicks, boxes, and scribbles, and ignores deep investigation into other prompt encoding like mask encoding.
Secondly, although our approach gives a probability estimation on an image to generate a prompt encoding vector to offer richer feedback cues, it 
inevitably introduces noise, which would affect performance improvement.
Thirdly, the existing performance by PVPUFormer on cross-domain learning is not so good, which is shown in the performance testing on medical datasets.

In the future, we intend to further integrate other modalities of interactive types, such as text, voice, \textit{etc.}, by utilizing the prompt encoding module to integrate different forms of user prompts.
Moreover, we will try to improve the probability estimation module of prompt encoding to reduce the noise information for performance improvement. Additionally, cross-domain or open-set scenarios have been challenging and prominent research topics for our future work.

\section{Conclusion} \label{Conclusion}
In this paper, we look into interactive image segmentation and propose a Probabilistic Visual Prompt Unified
Transformer (PVPUFormer) with effective unified visual prompt encoding. Beyond existing interactive segmentation methods, our approach deeply excavates the 
characteristics of diverse visual prompts and proposes a simple yet effective Probabilistic Prompt-unified Encoder (PPuE), which adopts a unified probabilistic representation to encode different prompts by considering both prompt and non-prompt cues in a probabilistic estimation way. To the best of our knowledge, this is the first probabilistic prompt encoding study, which could offer sufficient valuable feedback information for performance boosting. On this basis, our approach further introduces the Dual-cross Merging Attention (DMA) module and the Prompt-to-Pixel Contrastive (P$^2$C) loss to generate robust visual features, which is conductive to enhance the accuracy of mask prediction.  
Extensive experiments on a large number of natural and medical image datasets have been done, and the experimental results prove that the proposed components are effective for interactive image segmentation, yielding state-of-the-art performance as compared to the existing methods.

{\small
\bibliographystyle{IEEEtran}
\bibliography{refs}
}

\end{document}